\documentclass[ukenglish]{nik}
\usepackage[utf8]{inputenc}
\usepackage[T1]{fontenc}
\usepackage[english]{babel}
\usepackage{textcomp}
\usepackage{svg}
\usepackage{type1cm}
\usepackage{csquotes}
\usepackage{hyperref}

\usepackage{amsmath}
\usepackage{amssymb}
\usepackage{amsfonts}

\title{FlashRL: A Reinforcement Learning Platform for~Flash~Games}
\author{Per-Arne~Andersen \and Morten~Goodwin \and Ole-Christoffer~Granmo \\ University of Agder, Faculty of Engineering and Science\\ Serviceboks 509, NO-4898 Grimstad, Norway\\}
\date{}

\begin{document}

\maketitle

\begin{abstract}
    Reinforcement Learning (RL) is a research area that has blossomed tremendously in recent years and has shown remarkable potential in among others successfully playing computer games. However, there only exists a few game platforms that provide diversity in tasks and state-space needed to advance RL algorithms. The existing platforms offer  RL access to Atari- and a few web-based games, but no platform fully expose access to Flash games. This is unfortunate because applying RL to Flash games have potential to push the research of RL algorithms. 
    
    This paper introduces the Flash Reinforcement Learning platform (FlashRL) which attempts to fill this gap by providing an environment for thousands of Flash games on a novel platform for Flash automation. It opens up easy experimentation with RL algorithms for Flash games, which has previously been challenging. The platform shows excellent performance with as little as 5\% CPU utilization on consumer hardware. It shows promising results for novel reinforcement learning algorithms.

\end{abstract}


\newpage
\section{Introduction}

There are several challenges related to developing algorithms that can interact with human-level performance in real-world environments, such as computer games. Researchers often use toy experiments when working with \textit{Reinforcement Learning} (RL), because it is easier, cheaper and consumes less time to orchestrate. With several applications for RL in daily life, it has become an essential field of research \cite{Russell1995, Grost1989}. However, existing learning platforms for games have major limitations such as few game environments and little environment control.

\textit{OpenAI} is a non-profit company that is currently one of the leading researchers of RL. OpenAI \textit{Universe} is a software platform that has several game environments aimed at artificial research. The problem with this software is that individual developers are not directly permitted to supplement new environments to the repository, and there is little documentation on how to contribute to new environments. \textit{FlashRL} changes this with our proposed architecture as the control is given back to each researcher.

\textit{Adobe Flash} is a multimedia software platform used for the production of applications and animation. The Flash run-time was recently declared deprecated by Adobe, and by 2020, no longer supported. Flash is still frequently used in web applications, and there are several thousand games created for this platform. Several browsers have removed support for Flash, making it impossible to access the mentioned game environments. Games have proven to be an excellent area of machine learning benchmarking, due to size and diversity of its state-space. It is therefore essential to preserve Flash as an environment for reinforcement learning.

Automating Flash applications is a relatively untouched area. The technology has been succeeded by several better options for web development, for example, HTML5. This makes it hard for algorithms to control Flash environments programmatically. There are already reinforcement learning platforms that support Flash games as part of their game library, but these use browsers to execute the Flash run-time.

\begin{figure}[htp]
  \centering
  \includegraphics[width=1.0\textwidth]{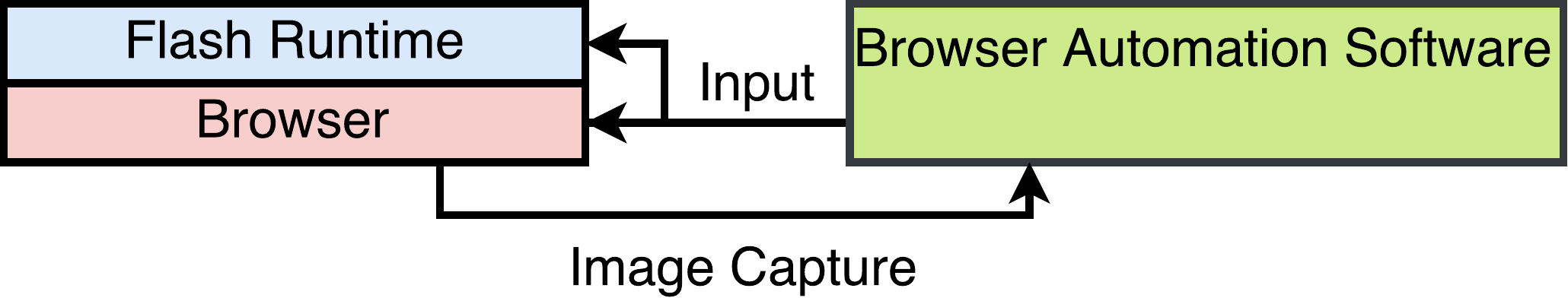}
  \caption{Interacting with Flash through browser automating}
  \label{fig:old_flash_interaction}
\end{figure}

Figure \ref{fig:old_flash_interaction} illustrates how interaction with the Flash environment would typically be carried out through browser automation software such as \textit{Selenium}. \textit{Selenium} can automate most modern browsers. It does not directly support Flash automation, but can easily be used for this purpose with minimal customisation \cite{Guru992017}. With the loss of browser support, the difficulty of controlling Flash applications increases, and there is a significant risk that excellent game environments for reinforcement learning are lost.
 
FlashRL is unique for reinforcement learning as it allows researchers to use any desired Flash environment. It gives full control of the game environment and is not based on running Flash applications in the browser. 

FlashRL is targeted research in reinforcement learning, but can also be used in other machine learning algorithms. It supports all kinds of Flash applications but is primarily used for agent-based gameplay. Several thousand game environments are included in the first release of the software\footnote{Author of this paper takes no credit for any game environments}. Multitask 2 is a Flash game that is excellent for reinforcement learning as it requires the agent to perform several tasks simultaneously. We show in this paper that our learning platform can be used to train novel reinforcement algorithms without any customisation.

In Section \ref{sec:related_work}, we discuss related work for existing learning platforms in machine learning. We also argue why web browsers are no longer viable as Flash run-time. Section \ref{sec:reinforcement_learning} briefly outline what reinforcement learning is and explains how Q-Learning works. Section \ref{sec:solution} outlines the proposed platform and thoroughly describe its underlying architecture. In Section \ref{sec:results} we show initial results of utilizing the proposed learning platform for reinforcement learning. At Section \ref{sec:conclusion} summarises the work and argue why the proposed learning platform is used for reinforcement learning research. Section \ref{sec:future_work} outlines a road-map for further development of the platform.

\section{Related Work}
\label{sec:related_work}
With the increasing popularity in RL, there is a need for flexible learning platforms. Several learning platforms exist that can run a limited number of games, but no platform that features an open-source interface with possibility to run \textit{any} Flash game.

Bellemare et al. provided in 2012 a learning platform \textit{Arcade Learning Environment} (ALE) that enabled scientists to conduct edge research in general deep learning \cite{Bellemare2015}. The package provided hundreds of Atari 2600 environments that in 2013 allowed Minh et al. to do a breakthrough with Deep Q-Learning and A3C. The platform has been a key component in several breakthroughs in RL research. \cite{Mnih2013, Mnih2015, Mnih2016}

In 2016, Brockman et al. from OpenAI released GYM which they referred to as \textit{"a toolkit for developing and comparing reinforcement learning algorithms"} \cite{Brockman2016}. GYM provides various types of environments from following technologies \cite{Brockman2016}: Algorithmic tasks, Atari 2600, Board games, Box2d physics engine, MuJoCo physics engine, and Text-based environments.
OpenAI also hosts a website where researchers can submit their performance for comparison between algorithms. GYM is open-source and encourages researchers to add support for their environments.

OpenAI recently released a new learning platform called \textit{Universe}. This environment further adds support for environments running inside VNC. It also supports running Flash games and browser applications. However, despite OpenAI's open-source policy, they do not allow researchers to add new environments to the repository. This limits the possibilities of running any environment. Universe is, however, a significant learning platform as it also has support for desktop games like Grand Theft Auto IV, that allow for research in autonomous driving \cite{Li2017}.

Selenium is a software for automating web browsers and is used primarily for unit-testing of web content. There were some efforts to create a version that allowed to interact with Flash content, but it was quickly abandoned. There is limited support for interacting with Flash, by selecting the DOM-Element in HTML and sending key-presses via Javascript. Several learning platforms utilize this method, but due to the deprecation of Flash in browsers, it is no longer a viable option.

\section{Reinforcement Learning}
\label{sec:reinforcement_learning}
Reinforcement learning can be considered hybrid between supervised and unsupervised learning. We implement what we call an agent that acts in our environment. This agent is placed in the unknown environment where it tries to maximize the environmental reward \cite{Sutton1998}.

Markov Decision Process (MDP) is a mathematical method of modeling decision-making within an environment. We often use this technique when utilizing model-based RL algorithms. In \textit{Q-Learning}, we do not try to model the MDP. Instead, we try to learn the optimal policy by estimating the action-value function $Q^*(s, a)$, yielding maximum expected reward in state s executing action a. The optimal policy can then be found by 

\begin{equation}
\pi(s) = argmax_aQ^*(s,a)
\end{equation}

This is derived from \textit{Bellman's Equation}, because we can consider $U(s) = max_aQ(s,a)$, the utility function to be true. This gives us the ability to derive following update-rule equation from Bellman's work:

\begin{equation}
\label{eq:qupdate}
Q(s,a) \leftarrow Q(s,a) + \underbrace{\alpha}_\text{LR} \Bigg( \underbrace{R(s)}_\text{Reward} + \underbrace{\gamma}_\text{Discount} \underbrace{max_{a^{'}} Q(s^{'},a^{'})}_\text{New Q} - \underbrace{Q(s,a)}_\text{Old Q}  \Bigg)
\end{equation}

This is an iterative process of propagating back the estimated Q-value for each discrete time-step in the environment. It is guaranteed to converge towards the optimal action-value function, $Q_i \rightarrow Q^*$ as i $\rightarrow \infty$ \cite{Sutton1998, Mnih2013}.
At the most basic level, Q-Learning utilize a table for storing $(s,a,r,s^{'})$ pairs. But we can instead use a non-linear function approximation in order to approximate $Q(s,a;\theta)$. $\theta$ describes tunable parameters for approximator. Artificial Neural Networks (ANN) are a popular function approximator, but training using ANN is relatively unstable.
%
%

\section{Flash Reinforcement Learning (FlashRL)}
\label{sec:solution}
The proposed platform is an interface that acts as a bridge between the \textit{Gnash Flash player} and the reinforcement learning algorithms. \textit{Flash Reinforcement Learning} (FlashRL) is a new platform that allows researchers to run algorithms on any Flash-based game efficiently.

\begin{figure}[htp]
  \centering
  \includegraphics[width=0.7\textwidth]{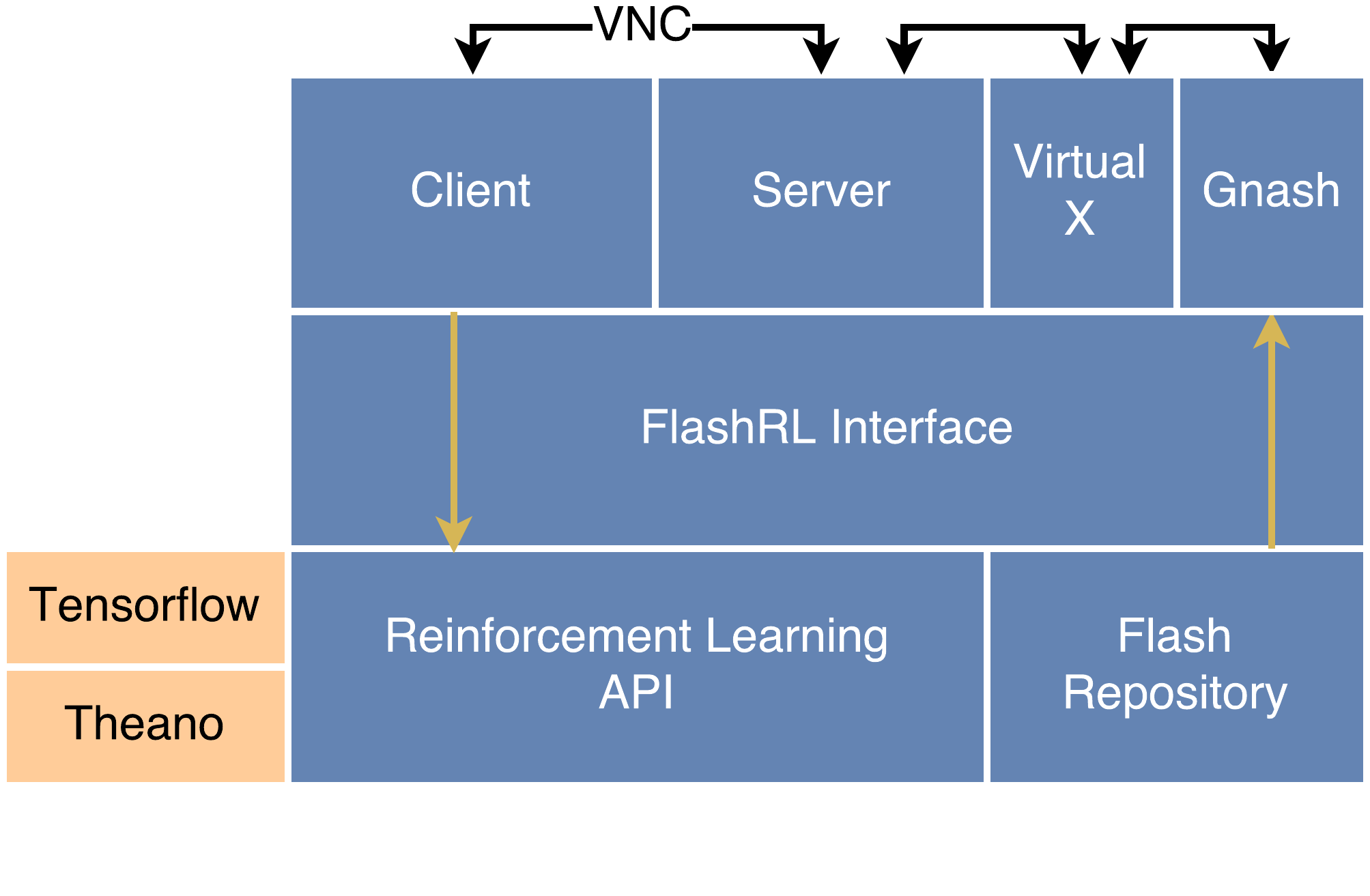}
  \caption{FlashRL Architecture Overview}
  \label{fig:flashrl_architecture}
\end{figure}
The learning platform is developed primarily for the operating system Linux but is likely to run on Cygwin with few modifications. There are several key components that FlashRL uses to operate adequate, see Figure \ref{fig:flashrl_architecture}. It uses a Linux library called XVFB to create a virtual frame-buffer that is used for graphics rendering \cite{Hunt2004}. Inside this frame-buffer, a Flash game chosen by the researcher is executed by a third party flash player, for example, \textit{Gnash}. A VNC server serves the XVFB frame-buffer and allows FlashRL to access it by utilizing a VNC Client. The VNC Client can then issue commands like keyboard presses and mouse movements.
The VNC Client \textit{pyVLC} was specially made for this learning platform. The code base originates from python-vnc-viewer \cite{Techtonik2015}. The last component of FlashRL is the Reinforcement Learning API that allows the developer to access the input/output of the VNC client. This makes it easy to develop sequenced algorithms by using the API callbacks or manually by threading.

\begin{figure}[htp]
  \centering
  \includegraphics[width=0.8\textwidth]{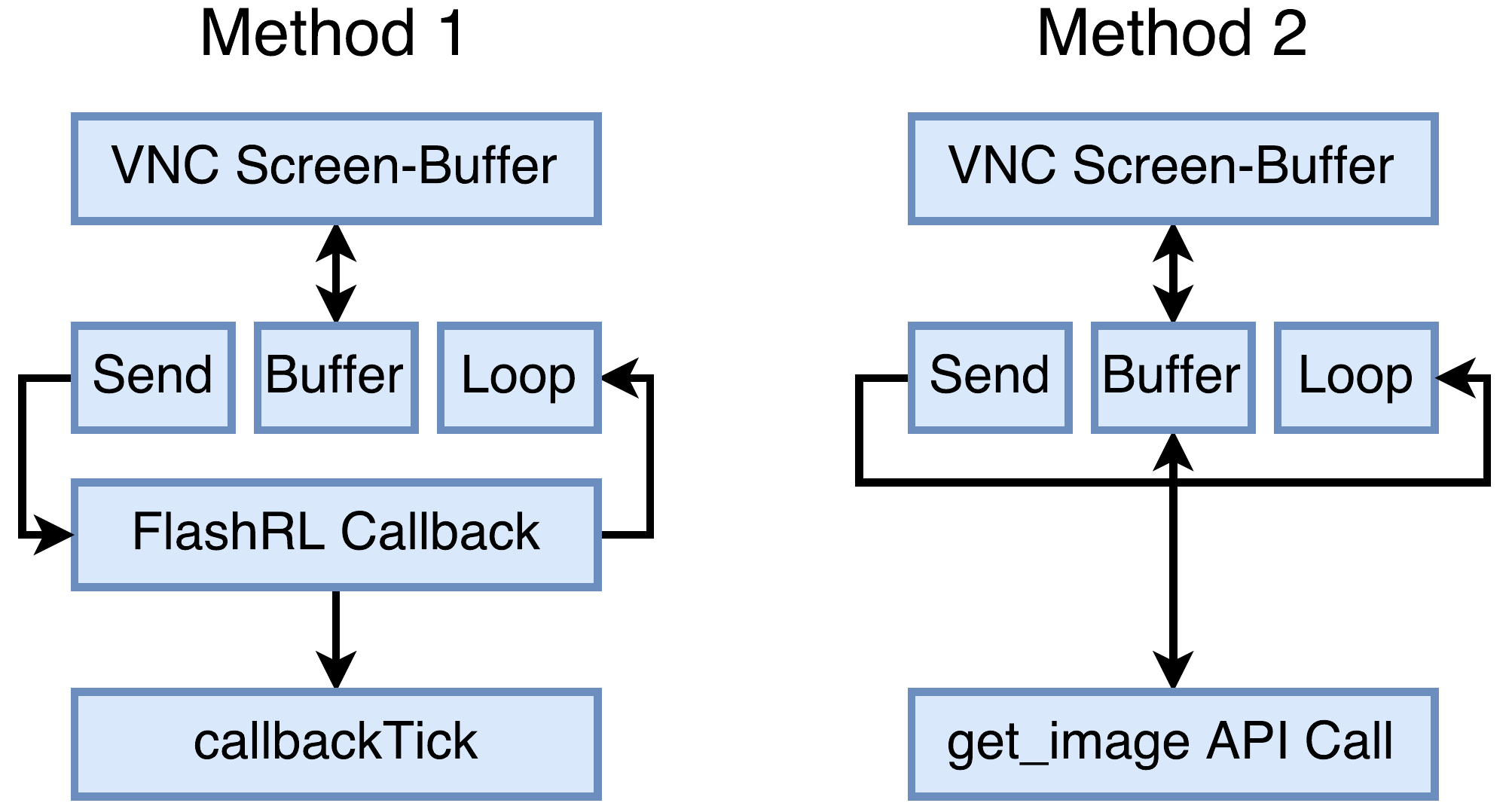}
  \caption{Frame-buffer Access Methods}
  \label{fig:execution_1}
\end{figure}

Figure \ref{fig:execution_1} illustrates two methods of accessing the frame-buffer from the Flash Game. Both approaches are sufficient to perform reinforcement learning, but each has its strength and weaknesses.
Method 1, seen in Figure \ref{fig:execution_1} allows the developer to get frames served at a fixed rate, for example, 60 frames per second. Method 2 does not restrict the frequency of how fast the frame-buffer is captured. This is preferable for developers that do not require images from fixed time-steps as it requires less processing power per frame. The framework was developed with deep learning in mind and is proven to work with Keras and Tensorflow.

\begin{figure}[htp]
    \centering
    \includegraphics[width=0.8\textwidth]{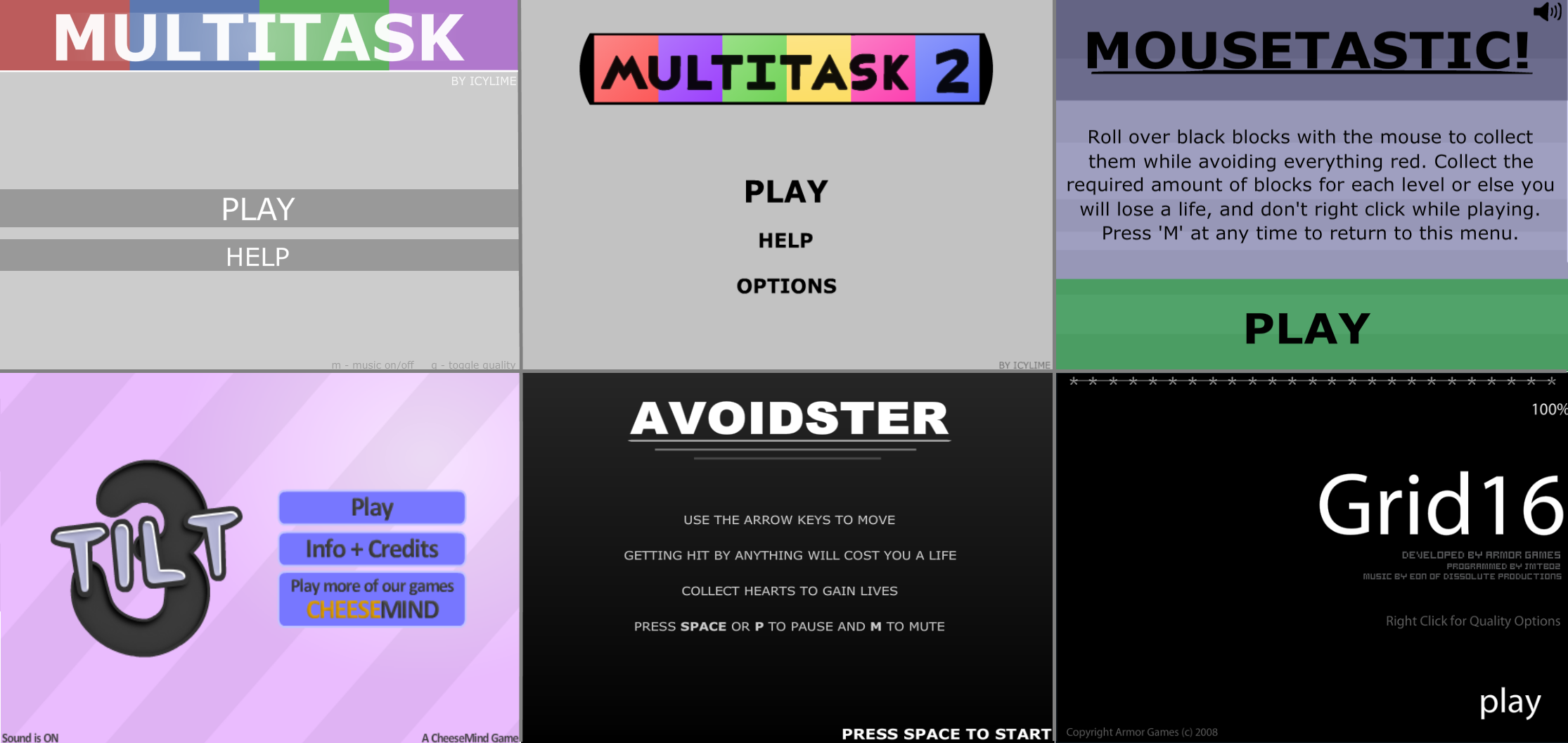}
    \caption{Selected environments from the FlashRL game repository}
    \label{fig:game_list}
\end{figure}

Several thousand game environments are shipped with the initial version of FlashRL. These game environments were gathered from different sources on the web. FlashRL has a relatively small code-base and to preserve this size, all of the Flash games are hosted remotely. The quality varies, and some of the games are not tested or labeled. Most games are however tested and can be played without issues, see Figure \ref{fig:game_list}. 

\section{Experiments}
\label{sec:results}
This section presents experiments of reinforcement learning algorithms applied in FlashRL. We use the game Multitask 2 \footnote{Multitask 2 - http://multitaskgames.com/multitask-2.html} to test the learning platform. Multitask 2 was chosen because it challenges the algorithm to master four different mini-games simultaneously.

The experiments are grouped in two. The first experiment determines the hardware requirements of the platform and benchmarks the speed of critical operations. The second experiment is an implementation of standard Deep Q-Learning trained on raw state images from Multitask 2 to perform game actions. The latter is meant as a proof of concept that RL algorithms can be applied in FlashRL.

All experiments were conducted on Ubuntu Linux 17.04 x64 running Python 3.5.3. The machine has 64GB memory, Nvidia GeForce 1080TI, and Intel I7-7770k as hardware.

\begin{figure}[htp]
    \centering
    \includegraphics[width=1.0\textwidth]{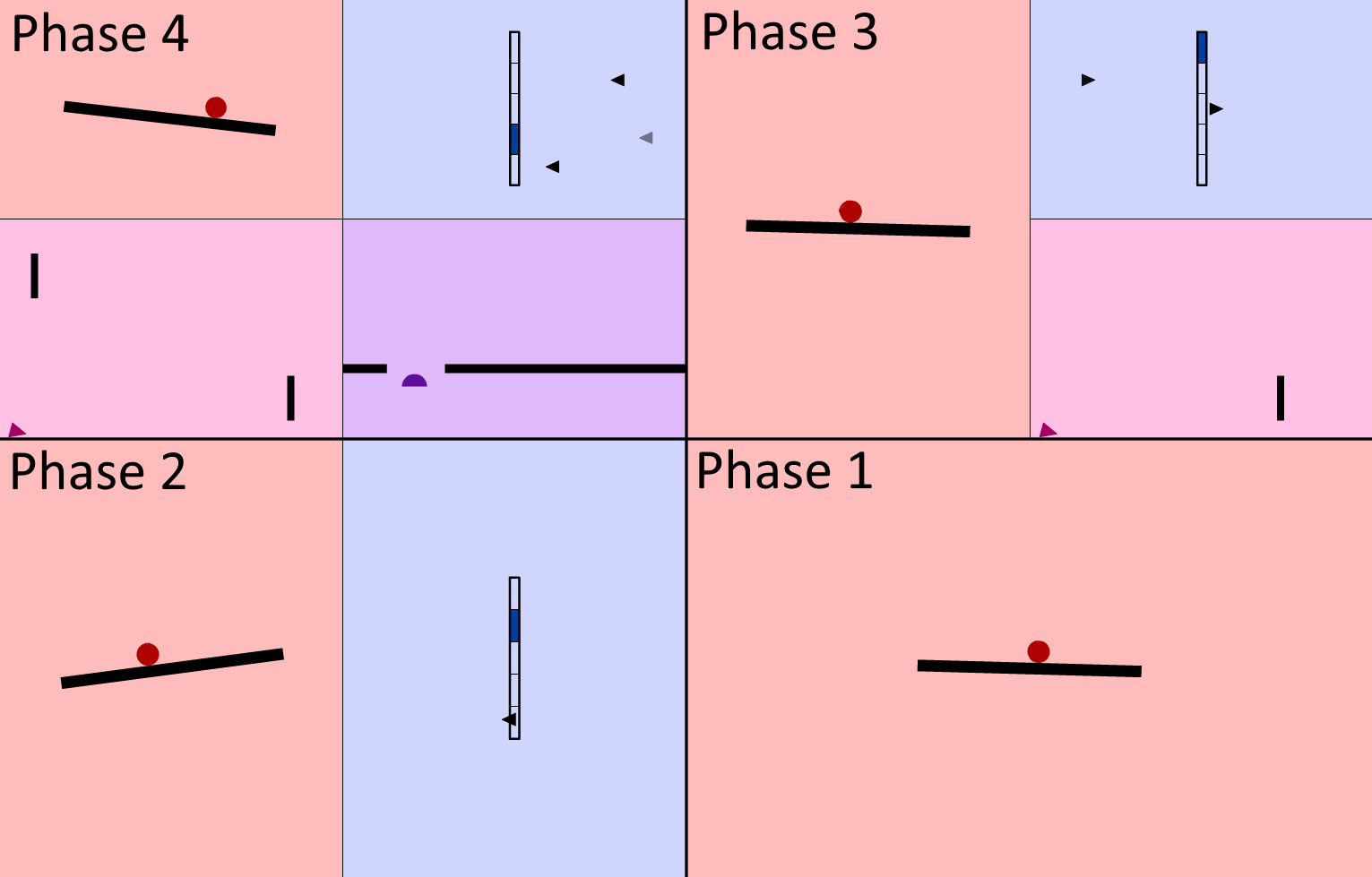}
    \caption{In-game footage of the game Multitask}
    \label{fig:multitask_game}
\end{figure}

\subsection{Multitask 2}\label{sec:multitask}

Figure \ref{fig:multitask_game} illustrates the game-play of Multitask 2. The game is split into four-game phases. The first phase (lower right corner in Figure \ref{fig:multitask_game}) is a single paddle that the player must balance a ball on. In state two (lower left corner in Figure \ref{fig:multitask_game}) , the player must control the second paddle to avoid arrows traveling towards it. The third phase (upper right corner in Figure \ref{fig:multitask_game}) consist of an arrow with mechanics relatable to the game Flappy Bird \cite{Piper2017}. In the final phase (upper left corner in Figure \ref{fig:multitask_game}), the player must additionally jump over holes on the ground.
For the player to succeed the game, he must control eight actions simultaneously. The score is calculated by adding a single point for each second survived in the game.

\subsection{Experiment 1: Hardware Requirements}
Recall from section \ref{sec:solution} that there are two methods of accessing the frame-buffer. The first method (Method 1) is based on retrieving the frame-buffer at fixed time intervals. The second method (Method 2) does not have any interval restriction. This makes Method 2 faster because it does not require sleep between frames. This causes the framework to consume all available CPU, which is not always preferable.

\begin{figure}[htp]
    \centering
    \includegraphics[width=1.0\textwidth]{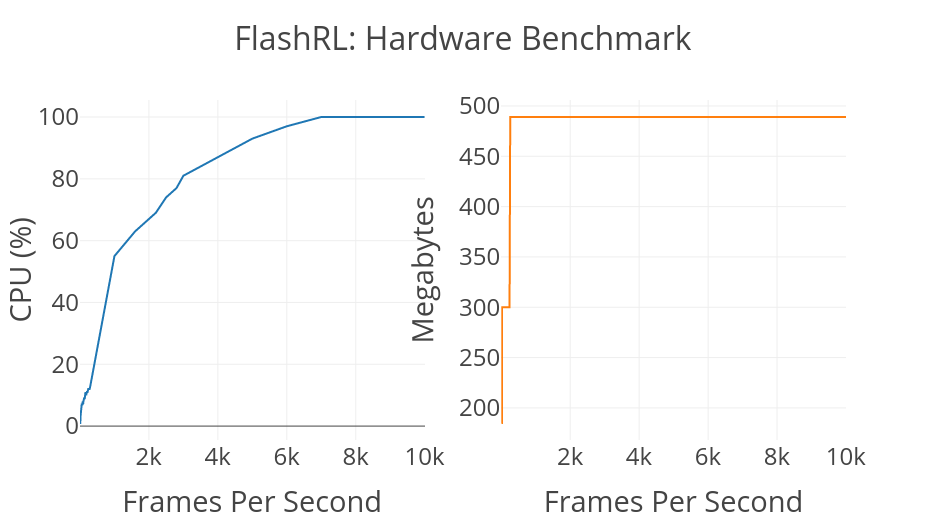}
    \caption{Hardware benchmark }
    \label{fig:hardware_benchmark}
\end{figure}

We can see from Figure \ref{fig:hardware_benchmark} that using Method 1 with the interval set to 30 fps uses approximately 5\% of the CPU. Increasing the interval to 300 increases it to 13\%. We gradually increased the interval until the CPU ran at maximum. A single I7-7700k can compute approximately 6300 fps images from the frame-buffer before struggling to keep up. 


The GPU Did not recognize any load during these test because the Flash environment is software rendered. Memory consumed were between 200MB and 500MB depending on the speed. We believe that the reason for memory increase is that Python does not garbage collect old frame-buffer snapshots between iterations, and therefore gets an increased memory load.
\subsection{Experiment 2: Reinforcement Learning}
Deep Q-Network (DQN) is a novel algorithm architecture developed by Minh et al. at Google DeepMind. It combines Q-Learning estimating Q-Values from a neural network. \cite{Mnih2013}

In our tests we used Double Q-Learning from Hasselt et al. \cite{VanHasselt2015}. We also used Dueling from Wang et al. that increases the learning precision by using two estimators: state-value and action-advantage function \cite{Wang2016}. We used a discount factor of 0.99, learning rate of 0.001 and mini-batch of 16. We used exploration/exploitation strategy with $\epsilon$-greedy where it started at 0.9 and finished at 0.1. The $\epsilon$ annealing was set to 10 000 steps. This is a relatively low epsilon phase. But it seemed to work well in this environment.

\begin{figure}[htp]
    \centering
    \includegraphics[width=1.0\textwidth]{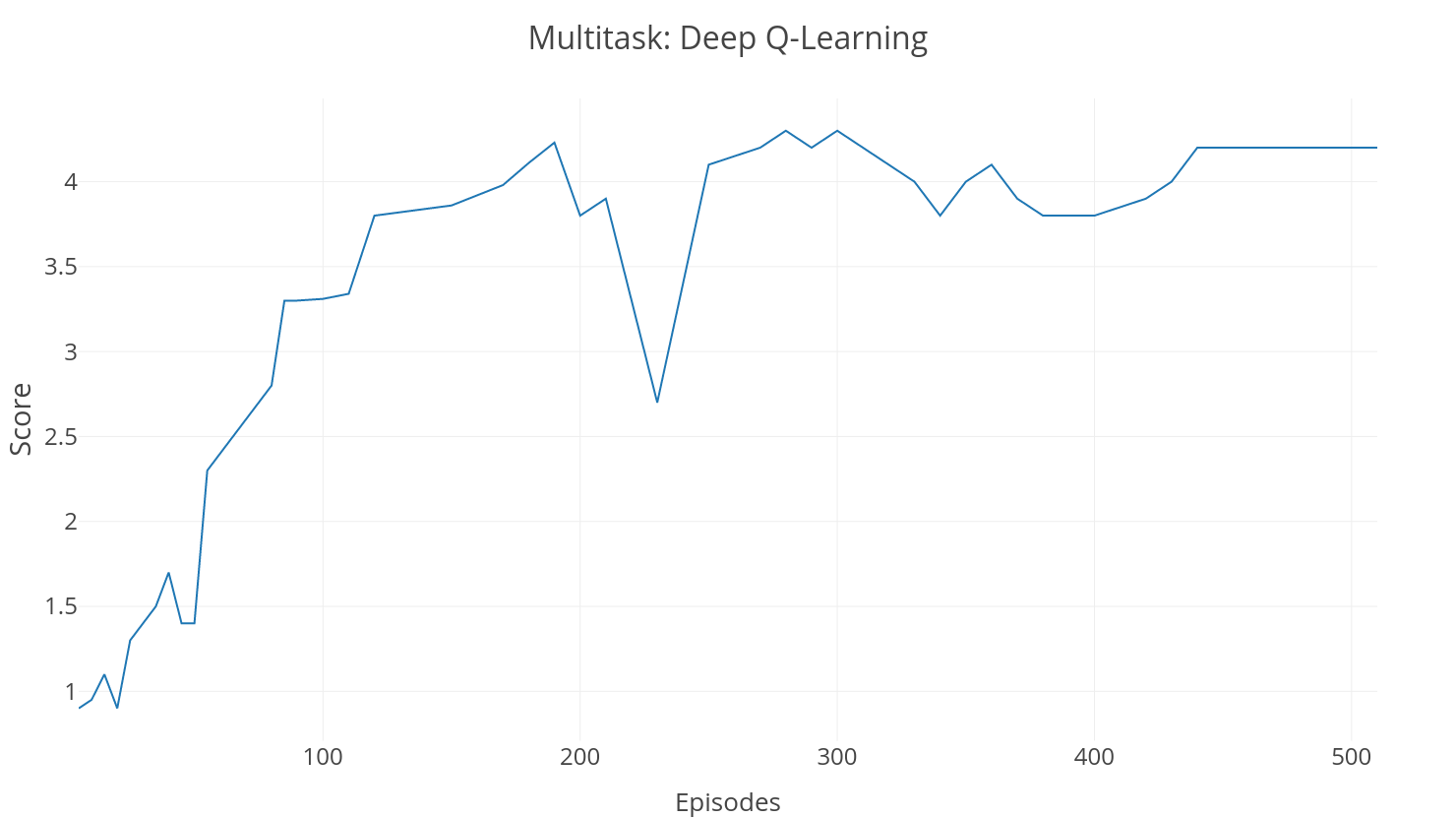}
    \caption{Deep Q-Learning Training}
    \label{fig:deep_q_learning}
\end{figure}

Figure \ref{fig:deep_q_learning} illustrates the training of DQN, where the x-axis represents episodes of the game and y-axis score before reaching the terminal state. The agent had troubles adapting to the third phase (see Section \ref{sec:multitask}). Phase 3 is relatively hard to master because it requires the user balance the arrow in the air. At around 230 episodes we saw a drop in score. This is because the network seems to prioritize the first phase of the game. It reached the second phase a few times but was not able to successfully control the paddle for longer periods of time. This is why it stales at approximately 400 episodes. We believe that the network could have performed better with additional training time. It trained for a total of two days. Hopefully, it will be easier to train the network when FlashRL can speed-forward games, see section \ref{sec:future_work}. The results are overall acceptable as we can see that FlashRL deliver quality states that a reinforcement learning agent can learn from.

\section{Conclusion}
\label{sec:conclusion}
FlashRL offers an easy-to-use architecture for performing RL in Flash-based games. It is demonstrated to work well for Multitask 2, one of the environments included. FlashRL fills the gap that emerged with the deprecation of Flash,   Its main focus is RL, but can also be used for other machine learning genres. This paper shows that FlashRL can be used to train RL algorithms, in particular, Multitask 2.  The work shows promising results and continuing to expand the game repository may provide new insights about RL in the future.

FlashRL will be kept alive as long as flash environments are an asset to the machine learning community. It is available to the public at \url{https://github.com/UIA-CAIR/FlashRL}, and can easily be adapted to every research requirement.


\section{Future Work}
\label{sec:future_work}
Several improvements are planned for FlashRL. This paper outlined features of the initial version of the FlashRL, and it is by far sufficient for simple reinforcement learning research. As seen in section \ref{sec:results}, a Deep Q-Learning based agent can successfully learn from the environment \textit{Multitask} and gradually perform better.

\subsection{Speed-forward Option}
Learning algorithms often require several thousand episodes to gain expert knowledge of the environment. FlashRL is currently limited to the speed of which the game loop is executed (usually 30 fps in real-time). An important improvement would be to lift this restriction and allow algorithms to train at an accelerated rate. This would certainly improve training duration of feedback based algorithms.

\subsection{Game Repository Analysis}
The game repository features many unlabeled, unrated and untested games. Some games are potentially useless in a machine learning setting and require a review. The review phase is time-consuming, and authors of this paper did not have enough time to analyze each of the environments manually. The goal is to add labels and categorize all games in the repository gradually.

\subsection{Website}
A future goal is to allow execution of algorithms from a web interface and to add gamification aspects to the library. This would potentially create competition between researchers much like Kaggle and OpenAI Universe.

\subsection{Cross-Platform Support}
FlashRL is in the initial version, only supported in Python 3 on the Linux platform. The goal is to extend it so that it also can run without modifications on Microsoft Windows operating systems.

\newpage
\renewcommand{\bibname}{References}
\bibliography{library}
\bibliographystyle{plain}
\end{document}